\documentclass[10pt,twocolumn,letterpaper]{article}
\usepackage[accsupp]{axessibility} 

\usepackage[pagenumbers]{wacv} 

\usepackage{graphicx}
\usepackage{booktabs}
\usepackage{textpos}
\usepackage{microtype}
\usepackage{marvosym}
\usepackage{amsmath, amssymb, amsthm, enumitem}
\usepackage[font=small]{caption}
\usepackage{algorithm}
\usepackage{algorithmic}
\usepackage{multirow}
\usepackage{soul}
\usepackage{xcolor}
\usepackage{newfloat}
\usepackage{listings}
\usepackage{pifont}
\usepackage{colortbl}
\usepackage{xfrac}
\usepackage{enumitem}
\usepackage{accents}
\usepackage{xspace}
\usepackage{pgfplots}
\pgfplotsset{compat=1.8}
\usepackage{graphicx}
\definecolor{beaublue}{rgb}{0.74, 0.83, 0.9}%

\usepackage{soul}

\usepackage{amsmath}
\DeclareMathOperator*{\argmax}{arg\,max}

\newcommand{\model}{\texttt{UNO}\xspace}
\newcommand{\bmodel}{\textbf{\texttt{UNO}}\xspace}

\definecolor{cvprblue}{rgb}{0.21,0.49,0.74}
\usepackage[pagebackref,breaklinks,colorlinks,allcolors=cvprblue]{hyperref}

\usepackage{graphicx}
\usepackage{blindtext}
\usepackage{natbib}
\usepackage{adjustbox}
\usepackage{subcaption}
\usepackage{caption}
\usepackage[font=small,labelfont=bf]{caption}
\usepackage{capt-of}
\usepackage{hhline}

\usepackage{multirow}
\usepackage{soul}
\usepackage[normalem]{ulem}
\usepackage{colortbl}
\usepackage{multicol}

\usepackage{multirow}

\definecolor{Gray}{gray}{0.85}
\definecolor{deeppink}{RGB}{255,20,147}
\definecolor{mygray}{gray}{0.95}
\definecolor{ggray}{RGB}{127,127,127}
\definecolor{aliceblue}{rgb}{0.94, 0.97, 1.0}

\usepackage{graphicx}
\usepackage{amssymb}
\usepackage{pifont}
\newcommand{\cmark}{\textcolor{OliveGreen}{\ding{51}}}
\newcommand{\xmark}{\textcolor{BrickRed}{\ding{55}}}

\newcommand{\myheading}[1]{\vspace{1ex}\noindent \textbf{#1}}

\newcommand{\pub}[1]{\color{gray}{\tiny{#1}}}

\def\wacvPaperID{3} 
\def\confName{WACV}
\def\confYear{2026}

\usepackage{graphicx}

\title{\texorpdfstring{\raisebox{-5pt}{\includegraphics[height=25pt,width=25pt]{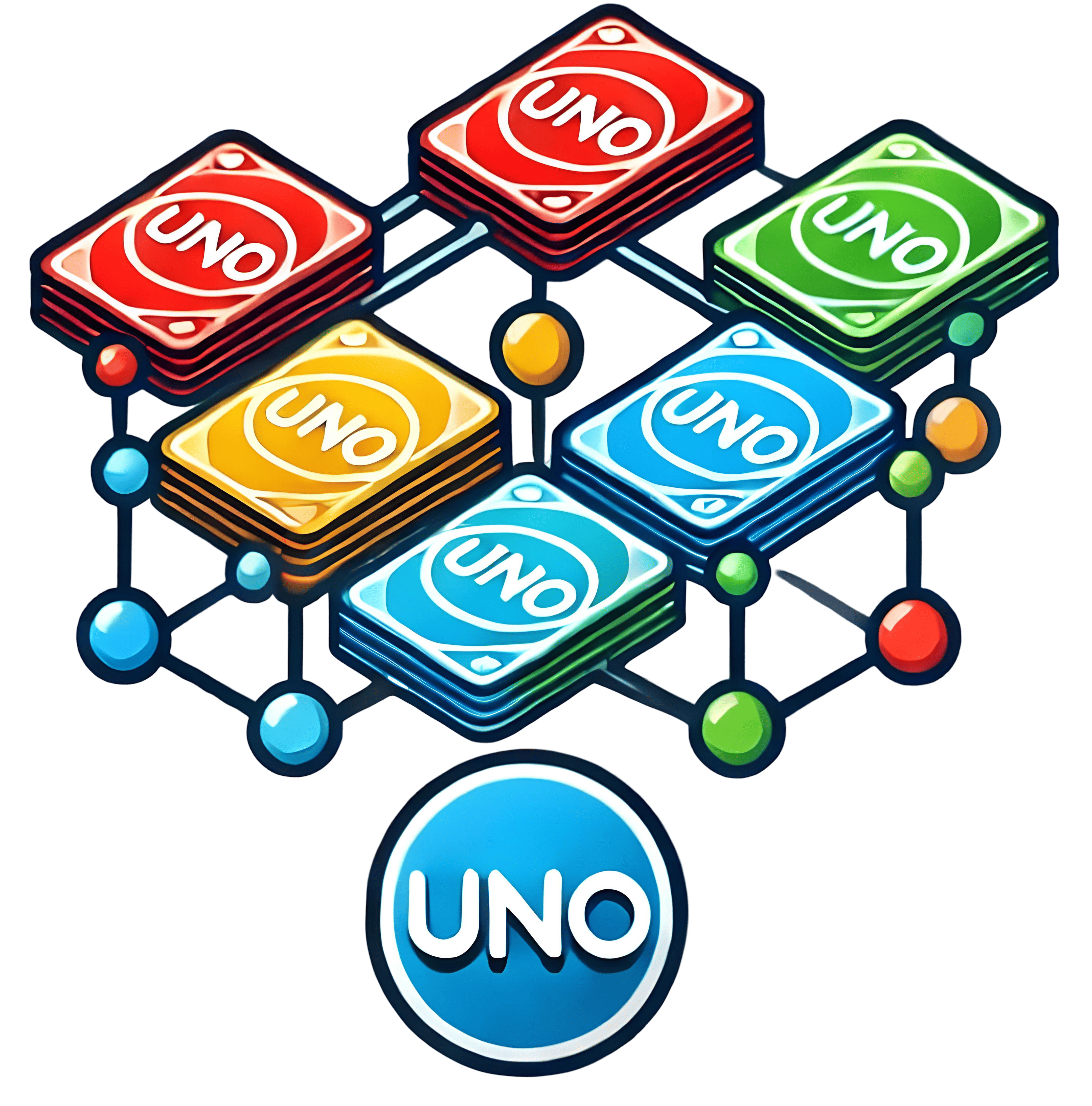}}}{Icon}\model: Unifying One-stage Video Scene Graph Generation\\via Object-Centric Visual Representation Learning}

\author{Huy Le$^{1}$\thanks{Corresponding author} \quad Nhat Chung$^1$ \quad Tung Kieu$^{2,3}$ \quad Jingkang Yang$^4$ \quad Ngan Le$^{5}$\\
$^1$FPT Software AI Center, Vietnam \quad
$^2$Aalborg University, Denmark \quad
$^3$Pioneer Centre for AI, Denmark \\
$^4$S-Lab, Nanyang Technological University, Singapore \quad
$^5$AICV Lab, University of Arkansas, USA \\
{\tt\small \{lehuy2316, nhatchung14\}@gmail.com} \quad
{\tt\small tungkvt@cs.aau.dk} \quad 
{\tt\small jingkang001@ntu.edu.sg} \quad
{\tt\small thile@uark.edu}
}

\begin{document}
        
\maketitle

\begin{abstract}
    Video Scene Graph Generation (VidSGG) aims to represent dynamic visual content by detecting objects and modeling their temporal interactions as structured graphs. Prior studies typically target either coarse-grained box-level or fine-grained panoptic pixel-level VidSGG, often requiring task-specific architectures and multi-stage training pipelines. In this paper, we present \model (UNified Object-centric VidSGG), a single-stage, unified framework that jointly addresses both tasks within an end-to-end architecture. \model is designed to minimize task-specific modifications and maximize parameter sharing, enabling generalization across different levels of visual granularity. The core of \model is an extended slot attention mechanism that decomposes visual features into object and relation slots. To ensure robust temporal modeling, we introduce object temporal consistency learning, which enforces consistent object representations across frames without relying on explicit tracking modules. Additionally, a dynamic triplet prediction module links relation slots to corresponding object pairs, capturing evolving interactions over time. We evaluate \model on standard box-level and pixel-level VidSGG benchmarks. Results demonstrate that \model not only achieves competitive performance across both tasks but also offers improved efficiency through a unified, object-centric design. The code is available at: \url{https://github.com/Fsoft-AIC/UNO}
\end{abstract}
\vspace{-1.5em}    
\section{Introduction}
\label{sec:intro}

\begin{figure}[!ht]
    \centering
    \includegraphics[width=1.00\linewidth]{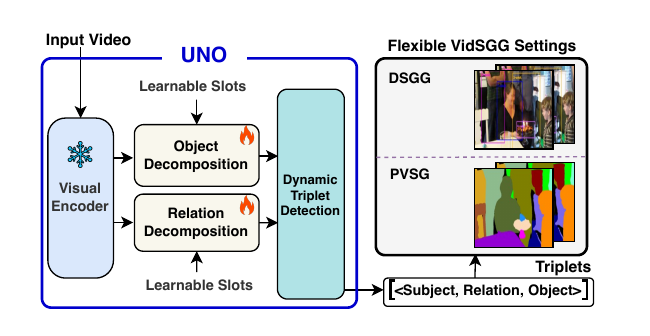}
    \vspace{-2em}
    \caption{\textbf{UNO.} We introduce \model, a unified framework for box-level VidSGG (DSGG) and pixel-level VidSGG (PVSG) settings.}
    \label{fig:teaser}
    \vspace{-18pt}
\end{figure}

Video Scene Graph Generation (VidSGG) aims to extract structured, dynamic representations from videos by modeling objects as nodes and their pairwise interactions as edges in spatio-temporal graphs. These structured representations offer both interpretability and compositionality, making VidSGG a critical component in various downstream tasks such as video understanding~\cite{ji2020action,av_scenegraph2024,yang2023pvsg,actionsg4ego2024,DBLP:conf/icassp/waver,DBLP:conf/mm/bima}, video reasoning~\cite{sc_actionreasoning2019,sganticipation2024} and robotic reasoning~\cite{shuran2020dynamic4robot}.

Current VidSGG research primarily follows two directions, distinguished by the level of visual granularity: box-level VidSGG (also referred to as Dynamic Scene Graph Generation or DSGG)~\cite{ji2020action,Wang_2024_CVPR,nag2023unbiased,li2022dynamic,teng2021target}, and panoptic pixel-level VidSGG (also known as Panoptic Video Scene Graph Generation or PVSG)~\cite{yang2023pvsg,panopticsg2024tpami,psgg4D_2023}. The former focuses on coarse-grained object representations using bounding boxes and typically models relationships at the frame level. The latter provides fine-grained, pixel-level representations using panoptic segmentation masks, where object trajectories are treated as graph nodes and interactions---including object-object and object-background---are captured throughout the video. Importantly, PVSG emphasizes the temporal consistency of object identities across frames. 

In scenarios requiring multi-level scene understanding, a unified VidSGG model capable of handling both tasks is highly desirable. Such a model could flexibly adapt to diverse visual representations and support a wider range of applications without task-specific architectural redesign. {However, achieving this unification is non-trivial due to the differing structural assumptions, temporal modeling requirements, and visual encoding strategies inherent to each task.} 
Prior attempts have relied on multi-stage pipelines involving either box-level or pixel-level representation, followed by a tracking module~\cite {li2022dynamic,wang2023cross,zhang2023end,yang2023pvsg, mcl2025aaai}, which introduce significant computational overhead and often result in suboptimal performance due to decoupled learning and limited parameter sharing.
%
On the other hand, designing a unified, end-to-end solution encounters the core challenge of learning a semantically consistent spatio-temporal representation that generalizes across varying levels of granularity while remaining aligned with the underlying video dynamics.

Beyond spatial granularity, VidSGG approaches also differ in their temporal-level representations, which can be broadly categorized into frame-level~\cite{cong2021spatial,li2022dynamic,nag2023unbiased,Wang_2024_CVPR} and tracklet-level~\cite{yang2023pvsg,mcl2025aaai} methods. Frame-level methods construct scene graphs independently at each frame, aligning naturally with box-level VidSGG, but often fall short in modeling long-term interactions and maintaining temporal consistency. In contrast, tracklet-level methods are more accurate because they link object instances across frames to capture temporal dynamics explicitly, a strategy more common in pixel-level VidSGG. 
This lack of temporal generality hinders their ability to capture coherent and continuous scene dynamics---especially in applications requiring both fine-grained relationship modeling and long-range temporal reasoning. 
A detailed comparison of \model with prior works is presented in Tab.~\ref{tab:benchmarkmethodlist}.

\begin{table}[t]
\centering
\caption{Comparison of different models on \textit{VidSGG tasks}.} 
\vspace{-1em}
\label{tab:benchmarkmethodlist}
 \resizebox{\linewidth}{!}{
 \begin{tabular}{l ccccc}
 \toprule
 \multicolumn{1}{l}{\multirow{2}{*}{\textbf{Methods}}} &  \multicolumn{1}{c}{\multirow{2}{*}{\textbf{One-stage}}} & \multicolumn{2}{c}{\textbf{Granularity-level}} & \multicolumn{2}{c}{\textbf{Temporal-level}} \\
 \cmidrule(lr){3-4}  \cmidrule(lr){5-6}
& & \multicolumn{1}{c}{\multirow{1}{*}{\textbf{Box}}} &
 \multicolumn{1}{c}{\multirow{1}{*}{\textbf{Pixel}}} & \multicolumn{1}{c}{\multirow{1}{*}{\textbf{Frame}}} &
 \multicolumn{1}{c}{\multirow{1}{*}{\textbf{Tracklet}}} \\
 \midrule


{STTran~\cite{cong2021spatial}}~\pub{ICCV'21} & \xmark & \cmark & \xmark & \cmark & \xmark \\

{APT~\cite{li2022dynamic}}~\pub{CVPR'22} & \xmark & \cmark & \xmark & \cmark & \xmark \\

{TEMPURA~\cite{nag2023unbiased}}~\pub{CVPR'23} & \xmark & \cmark & \xmark & \cmark & \xmark \\

{PVSG~\cite{yang2023pvsg}}~\pub{CVPR'23} & \xmark & \xmark & \cmark & \cmark & \cmark  \\

{OED~\cite{Wang_2024_CVPR}}~\pub{CVPR'24} & \cmark & \cmark & \xmark & \cmark & \xmark \\

{MCL~\cite{mcl2025aaai}}~\pub{AAAI'25} & \xmark & \xmark & \cmark & \cmark & \cmark  \\

{DIFFVSGG~\cite{DIFFVSGG}}~\pub{CVPR'25} & \xmark & \cmark & \xmark & \cmark & \xmark  \\

{VISA~\cite{visa2025}}~\pub{CVPR'25} & \xmark & \cmark & \xmark & \cmark & \xmark  \\
\midrule
\textbf{\model (Ours)} & \cmark & \cmark & \cmark & \cmark & \cmark \\
 \bottomrule
 \end{tabular}
 }
 \vspace{-1.5em}
\end{table}

To address this challenge, we propose \model, a unified, object-centric, single-stage
VidSGG framework that effectively supports both box-level and pixel-level tasks. Fig.~\ref{fig:teaser} illustrates the overall concept of \model. 
Our central hypothesis is that despite their differences, DSGG and PVSG share a common semantic context in object-centric representation which making it well-suited for such modeling.
At the heart of \model is an extended slot attention mechanism~\cite{locatello2020object} that decomposes visual feature maps into compact
object and relation slots. These slots serve as modular building blocks and form a shared latent representation space across both tasks. To ensure temporal consistency, we introduce \emph{object temporal consistency learning}, which enforces the alignment of object slots across frames without the need for explicit tracking. Furthermore, we propose a \emph{dynamic triplet prediction module} that efficiently associates relation slots with subject–object pairs while reducing redundancy in the predicted triplets.

We validate \model on two standard benchmarks: Action Genome~\cite{ji2020action} for DSGG task
and PVSG~\cite{yang2023pvsg} for PVSG task. Experimental results show that \model consistently outperforms state-of-the-art (SOTA) methods in both accuracy and computational efficiency. While \model builds on prior concepts such as slot attention and temporal contrastive consistency, our key contribution lies in recontextualizing and integrating them into a unified, one-stage framework specifically designed for VidSGG. To the best of our knowledge, \model is among the first approaches to jointly address both box-level and pixel-level VidSGG within an object-centric paradigm.
\section{Related Works}
\label{sec:related_works}



\myheading{Video Scene Graph Generation (VidSGG).}
\label{sec:vidsgg}
VidSGG is an extension of Scene Graph Generation~\cite{sggreview2023} that analyses videos to identify objects and their relationships, representing this information as a structured graph to support high-level video understanding tasks~\cite{av_scenegraph2024,actionsg4ego2024,sc_actionreasoning2019,sganticipation2024,sc_actionreg2021,shuran2020dynamic4robot}. 
Researchers have explored how VidSGG can be leveraged on different granularities of video content, from coarse bounding boxes~\cite{ji2020action,deng2019relation,lin2020gps,teng2021target, cong2021spatial} to fine-grained panoptic masks~\cite{yang2023pvsg,panopticsg2024tpami,psgg4D_2023}, to represent dynamic interactions among objects with varying levels of precision. In fact, the literature has largely diverged into two directions:  

\textit{Dynamic Scene Graph Generation--DSGG} ~\cite{ji2020action,deng2019relation,lin2020gps,teng2021target,cong2021spatial} adopts the box-level approach to VidSGG that involves detection and tracking of object instances, capturing both spatial and temporal relationships to form graphs. In particular, Action Genome~\cite{ji2020action} supports DSGG with bounding boxes, relationship labels, and actions of human-object interactions. Various strategies have been proposed to predict objects and classify their pair-wise relationships~\cite{deng2019relation, lin2020gps, cong2021spatial}. Recently, OED~\cite{Wang_2024_CVPR} reformulates DSGG as a set prediction problem on object boxes, and leverages pair-wise features to represent each subject-object pair within the scene graph. We refer to this as box-level VidSGG.

\textit{Video Panoptic Scene Graph Generation--PVSG
}~\cite{yang2023pvsg, panopticsg2024tpami, psgg4D_2023} requires nodes in scene graphs to be grounded by precise, pixel-level segmentation masks to facilitate fine-grained scene understanding. In the PVSG benchmark~\cite{yang2023pvsg}, frames in a video are assigned with panoptic masks that provide pixel-level detail of object and background boundaries, where methods~\cite{yang2023pvsg, panopticsg2024tpami} have had to capture both evolving object-background interactions and object-object interactions to create a cohesive, dynamic scene graph representation. We refer to this as pixel-level VidSGG.

Interestingly, VidSGG strategies are often multi-stage pipelines~\cite{deng2019relation,lin2020gps,teng2021target,cong2021spatial,panopticsg2024tpami,psgg4D_2023}, while OED~\cite{Wang_2024_CVPR} recently pioneered a one-stage VidSGG approach for DSGG, eliminating the need for external tracking or multi-stage optimization. Although OED could not be directly adopted into PVSG, it highlights an important design consideration for stable learning and efficient end-to-end VidSGG modeling. \emph{Building on this insight, our research introduces a novel one-stage framework that unifies dynamic scene graph generation for both DSGG and PVSG in videos, setting it apart from prior approaches.}

\myheading{Object-centric Representation Learning.}
\label{sec:slot}
Object-centric representation learning has been adopted to focus on entities~\cite{objectcentricprior4act2021, whatwhere2024, egoexoalign2023, sganticipation2024, actionslot2024} that are directly meaningful for study predictions. It is employed to uncover modular structures and independent mechanisms, such as objects and their relationships, from multi-object visual inputs~\cite{slotssm2024, rethinkimage2video2024, zhou2022slot}. 
Conventional models rely on object-/region-specific priors to facilitate reasoning and comprehension~\cite{objectcentricprior4act2021, viola2022, whatwhere2024, egoexoalign2023}. Meanwhile, recent works have leveraged slot attention~\cite{rethinkimage2video2024, actionslot2024, locatello2020object} to facilitate object-centric representations from raw scene features or for embodied and robotic features~\cite{slotvla,rethinkingoc}. In particular, slot attention has used GRU~\cite{chung2014empiricalgru} and competitive attention mechanisms to bind to modular structures in the input~\cite{slotssm2024, locatello2020object}, potentially maintaining them through time for video understanding~\cite{rethinkimage2video2024}. 
However, only a few studies have been done to consider an object-centric perspective for VidSGG besides straightforward visual detection, despite how it can consistently disentangle object semantics from general scene details. \emph{In this work, our research aims to extend their utility to capture objects' modular structures and relationships for a unified VidSGG.}


\begin{figure*}[!t]
    \centering
    \includegraphics[width=0.9\linewidth]{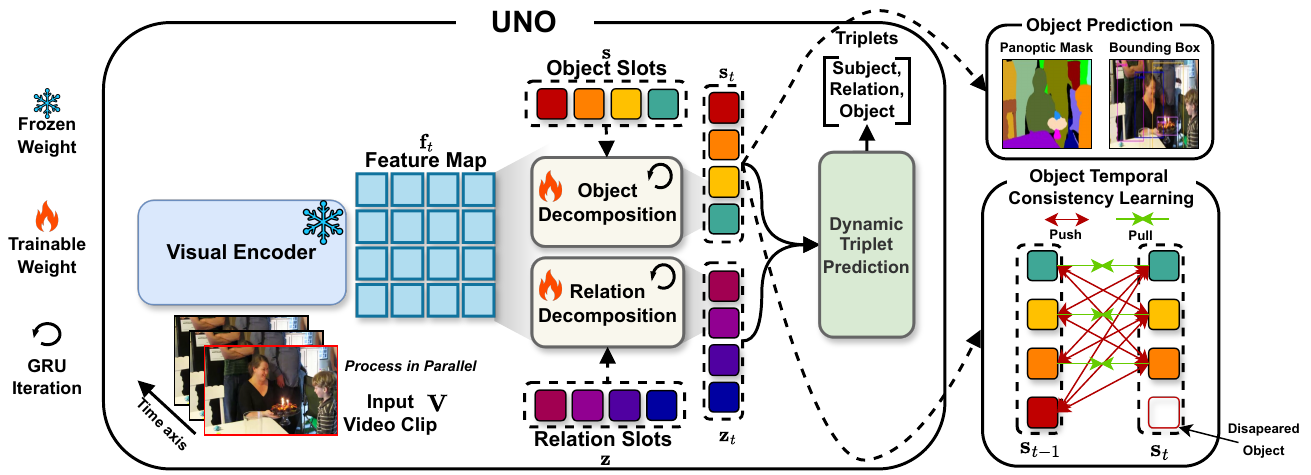}
    \vspace{-10pt}
    \caption{\textbf{UNO Framework.} Our architecture is powered with slot attention to efficiently decompose visual features into object and relation slots. The slots are also enabled with object temporal consistency learning to encourage their tracking through time. Finally, a dynamic triplet prediction module is integrated to align relation slots with their corresponding object slots, thereby obtaining the triplets of interest.} 
\label{fig:architecture}
\vspace{-15pt}
\end{figure*}

\section{Methodology}
\label{sec:method}

\subsection{Preliminary and Motivation}

In this subsection, we first introduce the definition and notation for existing VidSGG tasks, covering both DSGG and PVSG. We then explain how these tasks can be unified through our proposed one-stage framework, \model.

\myheading{Current VidSGG tasks.} 
Given a video $\mathbf{V} = \langle I_{1}, \ldots, I_{T} \rangle$ of $T$ frames, where each frame $I_{t} \in \mathbb{R}^{H_{\mathtt{in}} \times W_{\mathtt{in}} \times 3}$, VidSGG aims to produce a sequence of scene graphs $\mathbf{G} = \langle G_{1}, \ldots, G_{T} \rangle$, where each $G_{t}$ represents the scene graph for frame $I_{t}$. 
Each $G_{t}$ consists of triplets and is defined as $G_{t} = \{\textit{subject}, \textit{relation}, \textit{object}\}$.
The goal is to model the conditional probability $\mathbb{P}(\mathbf{G} \mid \mathbf{V})$. 
To simplify, existing methods~\cite{li2022dynamic} reformulate this as predicting detected objects and their pairwise relations.
\begin{equation}
    \small
    \mathbb{P}(\mathbf{G} \mid \mathbf{V}) = \mathbb{P}(\mathbf{B}, \mathbf{O}, \mathbf{R} \mid \mathbf{V}).
\end{equation}
\noindent Here, $\mathbf{B} = \langle \mathbf{B}_{1}, \ldots, \mathbf{B}_{T} \rangle$ is the set of bounding boxes for entire video $\mathbf{V}$
, where $\mathbf{B}_{t} = \{ B_1, \ldots, B_{M_{t}} \}$ is the set of bounding boxes of $M_{t}$ objects in the $t$-frame. Similarly, $\mathbf{O} = \langle \mathbf{O}_{1}, \ldots, \mathbf{O}_{T} \rangle$ is the set of object labels for entire video $\mathbf{V}$
, where $\mathbf{O}_{t} = \{ O_1, \ldots, O_{M_{t}} \}$ is the set of object labels 
for the $t$-frame. $\mathbf{R} = \langle \mathbf{R}_{1}, \ldots, \mathbf{R}_{T} \rangle$ is the set of relation for entire video $\mathbf{V}$
, where $\mathbf{R}_{t} = \{ R_1, \ldots, R_{L} \}$ is the set of relations for the $t$-frame. This approach is termed as coarse-grained box-level VidSGG task (DSGG~\cite{ji2020action}), and the conditional probability is factorized as follows:
\vspace{-0.5em}
\begin{equation}
    \small
    \label{E:dsgg_def}
    \begin{aligned}
        \mathbb{P}\left(\mathbf{G} \mid \mathbf{V}\right) & = \mathbb{P}\left(\mathbf{G}_{\text{DSGG}} \mid \mathbf{V}\right) \\
        & = \mathbb{P}\left(\mathbf{B} \mid \mathbf{V}\right)
        \mathbb{P}\left({\mathbf{O}} \mid \mathbf{B}, \mathbf{V}\right)
        \mathbb{P}\left(\mathbf{R} \mid \mathbf{O}, {\mathbf{B}}, \mathbf{V}\right). 
    \end{aligned}
\end{equation}
\noindent A prior study~\cite{yang2023pvsg} extended this formulation to handle the fine-grained pixel-level VidSGG by replacing the bounding boxes $\mathbf{B}$ with mask tubes $\mathbf{M}$, which represent each object in the entire video. This leads to the formulation: $\mathbb{P}(\mathbf{G} \mid \mathbf{V}) = \mathbb{P}(\mathbf{M}, \mathbf{O}, \mathbf{R} \mid \mathbf{V})$. We refer to this as the fine-grained pixel-level VidSGG task (PVSG~\cite{yang2023pvsg}), and the conditional probability is factorized as follows.
\begin{equation}
    \small
    \label{E:pvsg_def}
    \begin{aligned}    
        \mathbb{P}\left(\mathbf{G} \mid \mathbf{V}\right) &= \mathbb{P}\left(\mathbf{G}_{\text{PVSG}} \mid \mathbf{V}\right) \\
        &= \mathbb{P}\left(\mathbf{M} \mid \mathbf{V}\right)
        \mathbb{P}\left({\mathbf{O}} \mid \mathbf{M}, \mathbf{V}\right)
        \mathbb{P}\left(\mathbf{R} \mid \mathbf{O}, \mathbf{M}, \mathbf{V}\right).
    \end{aligned}
\end{equation}
\noindent where $\mathbf{M} = \langle{\mathbf{m}}_1, {\mathbf{m}}_2, \ldots, {\mathbf{m}}_{M_\mathtt{obj} + M_\mathtt{bg}}\rangle$ refers to the list of non-overlapping binary mask tubes of each object, where $M_\mathtt{obj}$ and $M_\mathtt{bg}$ is the number of objects and background appear in the video. For object $i$, the mask tube $\mathbf{m}_{i} \in\{0,1\}^{T \times H_\mathtt{in} \times W_\mathtt{in}}$ collects all tracked masks in each frame.

\myheading{Unifying VidSGG tasks.}
Given the current formulations of DSGG in Eq.\ref{E:dsgg_def} and PVSG in Eq.\ref{E:pvsg_def}, the primary distinction lies in their temporal granularity. DSGG operates at the frame level, where $\mathbf{O}_t$ denotes the set of objects detected within each individual frame, and thus does not enforce temporal consistency across frames. In contrast, PVSG leverages mask tubes, which inherently depend on maintaining object consistency throughout the video. 
A naïve approach to unifying VidSGG would result in a multi-stage pipeline of suboptimal performance and computational cost due to different modeling strategies in each task and stage.
To address this challenge, in this work, we consider two VidSGG tasks through a unified perspective and propose a framework to directly model $\mathbb{P}\left(\mathbf{G} \mid \mathbf{V}\right)$ such that {$\mathbf{G}=\{\mathbf{G_\text{DSGG}} \text{ or } \mathbf{G_\text{PVSG}}\}$} in an end-to-end training and inference manner. The primary challenge of our research lies in maintaining unified, spatio-temporal representations that align with video dynamics across both tasks, which we propose to address through an object-centric design, equipped with an object temporal consistency learning mechanism, ensuring a consistent and structured representation of objects and their relationships throughout the video sequence.

\myheading{Design Principles.}
{\model follows three key principles. First, it is a one-stage unified framework that minimizes task-specific modifications and multi-stage processing while maximizing parameter sharing through object-centric representation. Second, it establishes a strong baseline for diverse VidSGG tasks by reducing computational cost compared to multi-stage methods without sacrificing performance. Finally, instead of fusing bounding boxes and masks at the output level, \model employs a unified object-centric representation using Slot Attention, where object and relation slots serve as shared latent features and feed into task-specific heads for bounding box or panoptic mask prediction.}

\subsection{\bmodel Architecture}
An overview of \model is in Fig.~\ref{fig:architecture}.  
First, we utilize a frozen pre-trained visual encoder to extract feature maps from the last layer of each video frame.  
Next, we apply Slot Attention~\cite{locatello2020object} to decompose these feature maps into modular slots, effectively capturing both object and relation representations.
{These slots serve as shared latent features and are passed through a task-specific prediction head--either for bounding boxes or for panoptic masks.}
To ensure spatio-temporal consistency, we introduce object temporal consistency learning, reinforcing stable slot features across the video.  
Finally, we propose a dynamic triplet prediction mechanism that associates relation slots with their corresponding object pairs. Integrated into our end-to-end framework, this mechanism minimizes redundancy in triplet prediction while enhancing model efficiency.
\subsubsection{Visual Encoding}
First, we employ a frozen, pre-trained vision model as the Visual Encoder, where the last-layer feature map encodes rich object cues~\cite{ding2023betrayed,tian2023diffuse,caron2021emerging,simeoni2021localizing,oquab2023dinov2}, providing a spatial prior for learning object positions at varying granularities.  
Given a frame $I_t$, we extract its feature map $\mathbf{f}_{t} \in \mathbb{R}^{H_{\mathtt{enc}} \times W_{\mathtt{enc}} \times D_{\mathtt{enc}}}$, where $H_{\mathtt{enc}}$ and $W_{\mathtt{enc}}$ denote the spatial dimensions, and $D_{\mathtt{enc}}$ represents the feature channel size.  

Although pre-trained feature maps capture rich object information, they often entangle semantics, grouping similar objects~\cite{tian2023diffuse,ding2023betrayed}.  
To address this, a decomposition module is essential for disentangling these features into distinct semantics, enabling precise object and relation predictions for triplet construction.

\subsubsection{Object Decomposition}
We employ Slot Attention~\cite{locatello2020object}, inspired by recent advances in object discovery~\cite{scouter2021cvpr,zhou2022slot,actionslot2024}, as a clustering mechanism to group semantically meaningful patches from $\mathbf{f}_t$ into predefined slots, where each slot corresponds to a distinct object region.
Unlike previous studies~\cite{qian2023semantics,actionslot2024} that sample from a prior distribution, we initialize $N$ object slots (denoted as $\mathbf{s}$) as learnable tokens. We then decompose the feature map $\mathbf{f}_t$ of $t$-th frame into $N$ frame-wise object slot features $\mathbf{s}_t=\{\mathbf{s}_t^1, \ldots, \mathbf{s}_t^N\}$, $\mathbf{s}_t \in \mathbb{R}^{N\times D{\mathtt{slot}}}$.
This design encourages each slot to capture modular semantics~\cite{xu2022groupvit,DBLP:conf/iclr/JiaLH23}, enabling object-consistent decompositions through its unique \textit{competition mechanism} that supports maintaining coherent slot representation across frames. 

Formally, following the standard slot attention procedure~\cite{locatello2020object}, we employ three linear transformation heads to map the object slots $\mathbf{s}$ into \texttt{Query} $\mathbf{q}\in\mathbb{R}^{N\times D_{\mathtt{enc}}}$, while frame-wise feature maps $\mathbf{f}_t$ into \texttt{Key} $\mathbf{k}\in\mathbb{R}^{H_{\mathtt{enc}}W_{\mathtt{enc}}\times D_{\mathtt{enc}}}$, and \texttt{Value} $\mathbf{v}\in\mathbb{R}^{H_{\mathtt{enc}}W_{\mathtt{enc}}\times D_{\mathtt{enc}}}$.
We iteratively calculate attention score and update slot representations via Gated Recurrent Unit (GRU)~\cite{DBLP:conf/emnlp/ChoMGBBSB14}. Mathematically, we formulate each iteration as:

\vspace{-1.0em}
\begin{equation}
    \small
    \label{eq:slot_attention}
    \begin{aligned}
        &\Tilde{\mathbf{a}}_{i,j} = \frac{e^{\mathbf{a}_{i,j}}}{\sum_{l=1}^N e^{\mathbf{a}_{l,j}}}, \quad \text{where} \quad \mathbf{a} = \frac{1}{\sqrt{D}}\mathbf{q}\mathbf{k}^{\top}.\\
        &\mathbf{w}_{i,j} = \frac{\Tilde{\mathbf{a}}_{i,j}}{\sum_{l=1}^{H_{\mathtt{enc}}W_{\mathtt{enc}}}\Tilde{\mathbf{a}}_{i,l}}, \\
        &\mathbf{s}_t = \text{GRU}(\texttt{inputs}=\mathbf{wv}, \texttt{states}=\mathbf{s}_t)
\end{aligned}
\end{equation}
\vspace{-0.5em}

\noindent Here, the attention weights $\Tilde{\mathbf{a}}$ are normalized with $\mathit{softmax}$ along the slot dimension, and the weighted mean coefficient $\mathbf{w}$ aggregates the \texttt{Value} $\mathbf{v}$ to update the slots. 
This mechanism encourages competition among slots, ensuring each slot captures distinct object features. 
We use the object slots $\mathbf{s}_t$ at the final iteration as the distilled object tokens from the feature maps $\mathbf{f}_t$.

\myheading{Object Prediction.}  
Since each object slot captures multi-level granularity information, using a lightweight prediction head is sufficient for object class, bounding box, and mask prediction. 
Thus, we employ three lightweight prediction heads, each consisting of a feed-forward network (FFN) with two linear layers--a functional layer followed by a task-specific layer for each output. 
The classification head outputs object classes $\hat{\mathbf{O}}_{t} = \text{FFN}_{\texttt{cls}}(\mathbf{s}_t)$, while the box head predicts object coordinates $\hat{\mathbf{B}}_t = \text{FFN}_{\texttt{box}}(\mathbf{s}_t)$.
For mask prediction, a lightweight decoder with four transpose convolutions~\cite{DBLP:conf/cvpr/LongSD15} upsamples the feature map $\mathbf{f}_t$ to the original frame size, resulting in $\mathbf{f'}_{t} = \text{Dec}(\mathbf{f}_t)$.
The panoptic mask is then obtained by applying a matrix multiplication between the object slots $\mathbf{s}_{t}$ and the upsampled feature map $\mathbf{f'}_{t}$ in the mask head, resulting in a binary mask of each object slot $\hat{\mathbf{m}}_{t} = \text{FFN}_{\texttt{mask}}(\mathbf{s}_{t} \cdot \mathbf{f'}_{t})$.

\myheading{Object Temporal Consistency Learning.} 
Maintaining consistent spatio-temporal representations for object slots is crucial for tracking coherent object features over time.  
{Slot Attention provides a strong basis for object-centric video understanding. However, prior studies have noted that slot-based representations struggle to maintain temporal consistency across consecutive frames in a video~\cite{DBLP:conf/nips/ZadaianchukSM23}. 
To address this limitation, and inspired by prior works~\cite{DBLP:conf/cvpr/LiZPCCTL22,DBLP:journals/pami/FischerHPQCDY23}, we incorporate object temporal consistency learning using a contrastive loss.}
Specifically, slots matching the same ground truth index across frames are treated as positive samples, while all others are negatives.  
Formally, given the $i$-th object slot $\mathbf{s}^{i}_{t}$ at frame $t$, we define $\mathbf{s}^{i}_{t-1}$ as the positive target and $\mathbf{s}^{j}_{t-1}$ as negatives ($j \neq i$). The right side of Fig.~\ref{fig:architecture} illustrates the concept of object temporal consistency learning. The corresponding loss function is formulated as:

\vspace{-1.5em}
\begin{equation}
\small
\label{equ:consistency_loss}
    \mathcal{L}_{\text{consistency}} = - \sum_{\mathbf{s}_{t-1}^{i}}\log\frac{e^{(\mathbf{s}_{t}^{i}\cdot\mathbf{s}_{t-1}^{i})}}{e^{(\mathbf{s}_{t}^{i}\cdot\mathbf{s}_{t-1}^{i})} + \sum_{\mathbf{s}_{t-1}^{j}}e^{(\mathbf{s}_{t}^{i}\cdot\mathbf{s}_{t-1}^{j})}}.
\end{equation}
\vspace{-1.0em}

This loss encourages positive slots to stay close while pushing negative slots apart. By focusing only on matched slots, our method mitigates suboptimal updates from noisy, unmatched slots, ensuring that identical object slots remain aligned across frames. This alignment enables slots to refine each other's features, enhancing spatio-temporal consistency.

\subsubsection{Relation Decomposition}
Existing VidSGG approaches~\cite{Wang_2024_CVPR} predict relations sequentially after object prediction, leading to high computational complexity and potential inaccuracy if objects are missed. In contrast, \model predicts objects and relations simultaneously, enabling a parallelized execution that significantly reduces complexity.
Using Eq.~\ref{eq:slot_attention}, we apply an enhanced slot attention mechanism to decompose the feature map $\mathbf{f}_t$ into relation slots, where each slot captures potential interaction regions.
This mechanism enables relation slots to capture the entire spatial context of a frame, rather than being restricted to object pair intersections.
This broader coverage significantly enhances relation prediction performance and supports our dynamic triplet mechanism (see Sec.~\ref{sec:dynamic_prediction}).
Specifically, we initialize $K$ relation slots (denoted as $\mathbf{z}$) as learnable tokens. Then, for the $t$-th frame, we obtain frame-wise relation slot features $\mathbf{z}_t = \{\mathbf{z}_t^1, \ldots, \mathbf{z}_{t}^{K}\}$, where $\mathbf{z}_{t} \in \mathbb{R}^{K \times D{\mathtt{slot}}}$. Each frame-wise relation slot $\mathbf{z}_{t}$ is then passed through a classification FFN head to predict relations: $\hat{\mathbf{R}}_{t} = \text{FFN}_{\texttt{rel}}(\mathbf{z}_{t})$. Consequently, both relation slots $\mathbf{z}_{t}$ together with its relation class $\hat{\mathbf{R}}_{t}$ and object slots $\mathbf{s}_{t}$ are jointly extracted from $\mathbf{f}_{t}$ at time step $t$.

\begin{figure}[t]
\centering
\includegraphics[width=0.7\linewidth]{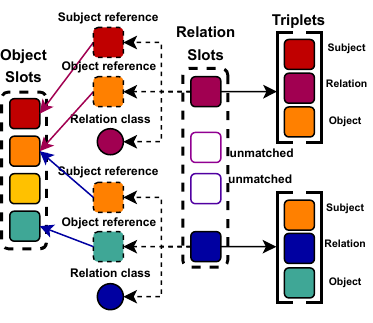}
\vspace{-10pt}
\caption{Dynamic triplet prediction module that predicts subject and object references to object slots from relation slots.}
\label{fig:dynamic_triplet}
\vspace{-10pt}
\end{figure}

\subsubsection{Dynamic Triplet Prediction}
\label{sec:dynamic_prediction}
Existing methods either learn an adjacency matrix~\cite{panopticsg2024tpami} to identify subject-object pairs or directly predict pairwise subject-object embeddings~\cite{Wang_2024_CVPR}, followed by sequential relation prediction.  
In contrast, our approach introduces a dynamic triplet prediction mechanism that directly associates $N$ objects with $K$ relations, eliminating the need to construct an $N \times N$ object pair matrix.  
{In theory, $K$ can be as large as $N^2$, since each object may interact with every other object. However, in practice, $K \ll N$ due to the inherent sparsity of real-world interactions—only a small subset of object pairs exhibit meaningful relationships. 
This reduces redundancy and duplication in triplet prediction while maintaining high performance. Fig.~\ref{fig:dynamic_triplet} illustrates the
proposed module. 

\myheading{Pairwise Index Matching.}
Each relation is defined as an interaction between two objects, with one as the subject and the other as the object.  
Thus, each relation slot inherently encodes information about the subject-object pair.  
We reformulate the problem as mapping relation slots to specialized representations that store the corresponding pair of object slots, enabling direct matching without constructing an $N \times N$ matrix.
Formally, given $K$ relation slots at the $t$-frame, $\mathbf{z}_t = \langle \mathbf{z}_{t}^{1}, \ldots, \mathbf{z}_{t}^{K} \rangle$, for each $\mathbf{z}_{t}^{j}$, we use two FFNs to generate subject-object pair of reference embeddings: $\mathbf{p}_{j}^{s}$ and $\mathbf{p}_{j}^{o}$.



\vspace{-1.5em}
\begin{equation}
    \small
    \mathbf{p}_{j}^{s} = \text{FFN}_s(\mathbf{z}_{t}^{j}), \quad \mathbf{p}_{j}^{o} = \text{FFN}_o(\mathbf{z}_{t}^{j})
\end{equation} 
\vspace{-1.0em}

Next, we find the indices of the subject and object that correspond to the object slots by matching the reference embeddings with the object slots $\mathbf{s}_{t}$ using a similarity function.
Specifically, we aim to identify the most relevant subject $\mathbf{s}^{i}_{t} \in \mathbf{s}_{t}$ and object $\mathbf{s}^{i'}_{t} \in \mathbf{s}_{t}$ given a specific subject reference embedding $\mathbf{p}_{j}^{s}$ and object reference embedding $\mathbf{p}_{j}^{o}$, respectively. This process is formulated as follows.

\vspace{-1.5em}
    \begin{equation} 
        \small
        \hat{i}_{j}^{s} = \argmax_{i \leq N}\:\text{sim}(\mathbf{p}_{j}^{s}, \mathbf{s}^{i}_{t}), \:
        \hat{i}_{j}^{o} = \argmax_{i' \leq N}\:\text{sim}(\mathbf{p}_{j}^{o}, \mathbf{s}^{i'}_{t})
    \end{equation}
\vspace{-1.0em}


\noindent Here, $\text{sim}(u, v) = \displaystyle\frac{u \cdot v}{||u|| \cdot |v||}$ is the similarity function. The predicted indices $\hat{i}_j^s$ and $\hat{i}_j^o$ correspond to the subject and object pair of the relation slot $\mathbf{z}_{t}^{j}$. This dynamic matching process effectively links relation slots with their most relevant pair of object slots, forming coherent triplets without relying on predefined adjacency matrices. This adaptive strategy enhances generalization across diverse object relations in videos.
The final triplet prediction of $t$-th frame is the set of $K$ triplets, $\{ \langle \mathbf{s}_t^{\hat{i}_j^s}, \mathbf{z}_t^i, \mathbf{s}_t^{\hat{i}_j^o} \rangle \}_{i=1}^K$, which can be replaced with corresponding bounding box and mask depends on the task.

\myheading{Triplet Duplication Reduction.} 
Unlike prior work~\cite{Wang_2024_CVPR}, which predicts triplets as a set and often results in duplicate detections, our method leverages slot attention with a built-in competitive mechanism to mitigate redundancy. This design ensures distinct object slots and relation slots focus on appropriate regions, producing unique object and relation predictions for triplets without requiring post-processing steps like Non-Max Suppression (NMS)~\cite{DBLP:conf/cvpr/sangBS17}.

\subsection{Training Objectives}
During training, we first perform Hungarian matching~\cite{DBLP:journals/jacm/EdmondsK72} between the predicted and ground-truth object boxes/masks to assign object slots, followed by supervision for detection and classification. The loss function for the DSGG task is defined as:
\vspace{-0.5em}
\begin{equation}
    \small
    \begin{aligned}
    \mathcal{L}_{\text{DSGG}} = & \lambda_{\text{obj\_cls}}\mathcal{L}_{\text{obj\_cls}}(\hat{\mathbf{O}}_t, \mathbf{O}_t)
    + \lambda_{\text{box}}\mathcal{L}_{\text{box}}(\hat{\mathbf{B}}_{t}, \mathbf{B}_{t}) \\
    & + \lambda_{\text{GIoU}}\mathcal{L}_{\text{GIoU}}(\hat{\mathbf{B}}_{t}, \mathbf{B}_{t}),
    \end{aligned}
\end{equation}
\noindent where $\mathcal{L}_{\text{obj\_cls}}$ is the Cross Entropy (CE) loss, $\mathcal{L}_{\text{box}}$ is the $\ell_{1}$ loss, and $\mathcal{L}_{\text{GIoU}}$ is the GIoU loss~\cite{DBLP:conf/cvpr/RezatofighiTGS019}. 

For the PVSG task, the loss is defined as:
\begin{equation}
    \small
    \begin{aligned}
    \mathcal{L}_{\text{PVSG}} = & \lambda_{\text{obj\_cls}}\mathcal{L}_{\text{obj\_cls}}(\hat{\mathbf{O}}_{t}, \mathbf{O}_{t})
     + \lambda_{\text{mask}}\mathcal{L}_{\text{mask}}(\hat{\mathbf{m}}_{t}, \mathbf{m}_{t}) \\
    & + \lambda_{\text{dice}}\mathcal{L}_{\text{dice}}(\hat{\mathbf{m}}_{t}, \mathbf{m}_{t}),
    \end{aligned}
\end{equation}
\noindent where $\mathcal{L}_{\text{mask}}$ is the CE loss, and $\mathcal{L}_{\text{dice}}$ is the Dice loss~\cite{DBLP:conf/miccai/SudreLVOC17,DBLP:conf/eccv/WangKSJL20}.

Finally, we re-apply Hungarian matching between the predicted and ground-truth relations to align relation slots while supervising both relation classification and index prediction. The matching between the indices of the predicted subjects/objects and their ground-truth indices is formulated as a classification problem, where indices are converted to one-hot vectors. The relation loss is formulated as follows:
\begin{equation}
    \small
    \begin{aligned}
    \mathcal{L}_{\text{Rel}} = & \lambda_{\text{rel\_cls}}\mathcal{L}_{\text{rel\_cls}}(\hat{\mathbf{R}}_t, \mathbf{R}_t) 
    + \lambda_{\text{sidx}}\mathcal{L}_{\text{sidx}}(\hat{i}^s, i^s) \\
    & + \lambda_{\text{oidx}}\mathcal{L}_{\text{oidx}}(\hat{i}^o, i^o),
    \end{aligned}
\end{equation}
\noindent where $\mathcal{L}_{\text{rel\_cls}}$, $\mathcal{L}_{\text{sidx}}$, and $\mathcal{L}_{\text{oidx}}$ are all CE losses.

\begin{table*}[htbp]
\centering
\caption{Performance comparison with SOTA DSGG methods on Action Genome dataset. The best results are in \textbf{bold}.}
\vspace{-1em}
\label{tab:sota_dsgg}
\setlength{\tabcolsep}{5pt}
\resizebox{\linewidth}{!}{
\begin{tabular}{l c cccccc cccccc}
\toprule
 & & \multicolumn{6}{c}{\textbf{With Constraint}} & \multicolumn{6}{c}{\textbf{No Constraint}}\\
 \cmidrule(lr){3-8} \cmidrule(lr){9-14}
 \multicolumn{1}{c}{\textbf{Method}} & \multicolumn{1}{c}{\textbf{Backbone}} &
   \multicolumn{3}{c}{\textbf{SGDET}}  &  \multicolumn{3}{c}{\textbf{PredCLS}} & \multicolumn{3}{c}{\textbf{SGDET}}  & \multicolumn{3}{c}{\textbf{PredCLS}}\\ 
   \cmidrule(lr){3-5} \cmidrule(lr){6-8} \cmidrule(lr){9-11} \cmidrule(lr){12-14} 
    & & R@10$\uparrow$ & R@20$\uparrow$ & R@50$\uparrow$ & R@10$\uparrow$ & R@20$\uparrow$ & R@50$\uparrow$ & R@10$\uparrow$ & R@20$\uparrow$ & R@50$\uparrow$ & R@10$\uparrow$ & R@20$\uparrow$ & R@50$\uparrow$ \\
\midrule
\multicolumn{14}{l}{\cellcolor{gray!30}\textbf{Multi-stage Method}} \\

{STTran~\cite{cong2021spatial}}~\pub{ICCV'21} & ResNet-101 &25.2 &34.1 &37.0 &68.6 &71.8 &71.8 &24.6 &36.2 &48.8 &77.9 &94.2 &99.1 \\

{APT~\cite{li2022dynamic}}~\pub{CVPR'22} & ResNet-101 & 26.3 & {36.1} & {38.3} &69.4 &{73.8} & {73.8} &25.7 &37.9 &50.1 &78.5 &95.1 &99.2 \\

{STTran-TPI~\cite{wang2022dynamic}}~\pub{ACM MM'22} & ResNet-101 & 26.2 &34.6 &37.4 &69.7 &72.6 &72.6 &- &- &- &- &- &- \\

{TR\textsuperscript{2}~\cite{wang2023cross}}~\pub{ICRA'23} & ResNet-101 &26.8 &35.5 &{38.3} &{70.9} &{73.8} &{73.8} &27.8 &39.2 &50.0 &83.1 &{96.6} &{99.9} \\

{VsCGG~\cite{DBLP:conf/mm/LuCSLWH23}}~\pub{ACM MM'23} & ResNet-101 & 27.4 & 35.8 & 38.2 & 70.1 & 73.4 & 73.5 &29.3 & 40.2 & 48.9 & 78.8 & 94.9 & 99.2 \\

{TEMPURA~\cite{nag2023unbiased}}~\pub{CVPR'23} & ResNet-101 &28.1 &33.4 &34.9 &68.8 &71.5 &71.5 &29.8 &38.1 &46.4 &80.4 &94.2 &99.4 \\

{DSG-DETR~\cite{feng2023exploiting}}~\pub{WACV'23} & ResNet-101 & {30.3} &34.8 &36.1 &- &- &- &{32.1} &{40.9} &48.3  &- &- &- \\

{TPT~\cite{zhang2023end}}~\pub{TMM'23} & ResNet-101 &- &- &- &- &- &- &32.0 &39.6 &{51.5} & {85.6} &{97.4} &{99.9} \\

{TD$^\text{2}$-Net~\cite{DBLP:conf/aaai/LinSZYWT24}}~\pub{AAAI'24} & ResNet-101 & 28.7 & - & 37.1 & 70.1 & - & 73.1 & 30.5 & - & 49.3 &81.7 & - & 99.8\\

\multicolumn{14}{l}{\cellcolor{gray!30}\textbf{One-stage Method}} \\

{OED{~\cite{Wang_2024_CVPR}}~\pub{CVPR'24}}
& ResNet-50 & {33.5} & {40.9} & {48.9}  & {73.0} & {76.1} & {76.1} & {35.3} & {44.0} & {51.8} &{83.3} &95.3 &99.2 \\

\midrule
\rowcolor{aliceblue}
& ResNet-50 & {35.4} &{42.2} &{49.5}  &{73.7} &{76.9} &{78.1} &{36.6} &{46.1} &{53.9} &{84.7} &{96.1} &{99.9} \\
\rowcolor{aliceblue}
& ViT-S/14 & {36.7} &{43.1} &{50.2}  &{74.2} &{78.5} &{79.6} &{37.5} &{47.5} &{54.5} &{85.9} &{96.6} &\textbf{100.0} \\
\rowcolor{aliceblue}
& ViT-B/14 & {38.2} &{44.7} &{51.9}  &{75.6} &{79.4} &{80.4} &{39.9} &{48.2} &{56.3} &{87.4} &{97.2} &\textbf{100.0} \\
\rowcolor{aliceblue}
\multirow{-4}{*}{\textbf{\model (Ours)}}& ViT-L/14 & \textbf{39.3} &\textbf{45.2} &\textbf{53.8}  &\textbf{76.8} &\textbf{80.3} &\textbf{82.5} &\textbf{40.8} &\textbf{49.7} &\textbf{57.1} &\textbf{88.3} &\textbf{98.1} &\bf{100.0}\\

\bottomrule
\end{tabular}}
\vspace{-10pt}
\end{table*}
\begin{table*}[htbp]
\centering
\caption{Performance comparison with SOTA PVSG methods on PVSG dataset. The best results are in \textbf{bold}. Next, $\diamondsuit$ and $\heartsuit$ stands for the relation predictor~\cite{yang2023pvsg}: 1D Convolution and Transformer Encoder, respectively.}
\vspace{-8pt}
\setlength{\tabcolsep}{4pt}
\resizebox{\linewidth}{!}{
\begin{tabular}{l c cccccc cccccc}
\toprule
     \multicolumn{1}{c}{\multirow{3}{*}{\textbf{Method}}} &
     \multicolumn{1}{c}{\multirow{3}{*}{\textbf{Backbone}}} &
     \multicolumn{6}{c}{\textbf{vIOU Threshold = 0.5}} & \multicolumn{6}{c}{\textbf{vIOU Threshold = 0.1}}\\
  \cmidrule(lr){3-8} \cmidrule(lr){9-14}
   & & R@20↑ & R@50↑ & R@100↑ & mR@20↑ & mR@50↑ & mR@100↑ & R@20↑ & R@50↑ & R@100↑ & mR@20↑ & mR@50↑ & mR@100↑ \\
\midrule
\rowcolor{ggray!30}\multicolumn{14}{l}{\textbf{Multi-stage Method}} \\

\multicolumn{14}{c}{\textbf{Image Panoptic Segmentation + Tracking}~\cite{cheng2021mask2former,wang2021different}} \\
{$\diamondsuit $PVSG~\cite{yang2023pvsg}} & ResNet-50 & {3.88} & 5.24 & {6.71} & 2.55 & 3.29 & {5.36} & {10.06} & {14.99} & {18.13} & {8.98} &  {12.21} & {15.47} \\
{$\heartsuit $PVSG~\cite{yang2023pvsg}} & ResNet-50 & {3.88} & {5.66} & 6.18 & {2.81} & {4.12} & 4.44 & 9.01 & 14.88 & 17.51 & 6.69 & 11.28 & 13.20 \\
{$\heartsuit$MCL~\cite{mcl2025aaai}} & ResNet-50 & 3.98 & 5.97 & 7.44 & 2.98 & 4.20 & 5.15 & 10.59 & 16.98 & 22.33 & 9.56 & 12.39 & 17.47 \\
{$\diamondsuit$MCL~\cite{mcl2025aaai}} & ResNet-50 & {4.51} & {6.08} & {7.76} & {3.56} & {4.38} & {5.86} & {11.43} & {17.30} & {22.85} & {9.57} & {13.13} & {17.48} \\
\midrule
\multicolumn{14}{c}{\textbf{Video Panoptic Segmentation}~\cite{cheng2021mask2former,li2022videoknet}} \\
{$\diamondsuit $PVSG~\cite{yang2023pvsg}} & ResNet-50 & {0.42} & 0.63 & 0.63 & 0.25 & 0.67 & 0.67 & {8.07} & {11.01} & {12.89} & {7.84} & {9.78} & {10.77} \\
{$\heartsuit $PVSG~\cite{yang2023pvsg}} & ResNet-50 & {0.42} & {0.73} & {1.05} & {0.61} & {0.76} & {0.92} & 6.50 & 9.64 & 12.26  & 5.75 & 8.25 & 9.51 \\
{$\heartsuit$MCL~\cite{mcl2025aaai}} & ResNet-50 & 0.63 & 1.05 & 1.05 & 0.83 & 0.76 & 0.76 & 6.71 & 10.27 & 13.42 & 6.94 & 8.68 & 12.09 \\
{$\diamondsuit$MCL~\cite{mcl2025aaai}} & ResNet-50 & 0.84 & 1.26 & 1.26 & 0.98  & 1.22 & 1.22 & 8.18 & 12.90 & 14.22 & 8.00 & 11.47 & 13.59 \\
\rowcolor{ggray!30}\multicolumn{14}{l}{\textbf{One-stage Method}} \\
\rowcolor{aliceblue} & ResNet-50 & 
{6.23} &{7.37} &{8.65} &{5.60} &{6.84} &{8.21} &{13.83} &{19.27} &{24.63} &{11.65} &{15.94} &{19.99} \\
\rowcolor{aliceblue} {\textbf{\model (Ours)}} & ViT-S/14 & 
{7.45} &{8.46} &{9.69} &{6.83} &{7.50} &{9.26} &{14.11} &{20.71} &{25.11} &{12.40} &{16.82} &{20.78} \\
\rowcolor{aliceblue} & ViT-B/14 & 
{8.71} &{9.19} &{10.46} &{7.56} &{8.99} &{9.83} & {15.76} &{21.87} &{26.58} &{13.12} &{17.25} &{21.81} \\
\rowcolor{aliceblue} & ViT-L/14 & \textbf{9.44} & \textbf{10.83} &\textbf{11.59} &\textbf{8.25} &\textbf{9.72} &\textbf{10.86} & \textbf{17.54} &\textbf{23.82} &\textbf{27.32} &\textbf{14.81} &\textbf{18.31} &\bf{22.14} \\
\bottomrule
\end{tabular}}
\label{tab:pvsg_sota}
\vspace{-15pt}
\end{table*}

\begin{table}[!t]
\footnotesize
\centering
\caption{{Multi-task training ablation on PVSG dataset.}}
\vspace{-1em}
\label{tab:multi_task}
 \setlength{\tabcolsep}{2pt}
\begin{tabular}{lccccc}
\toprule
{{Training setting}} &
{AP50$\uparrow$} & {PQ$\uparrow$} & {R@20$\uparrow$} \\ 
\midrule

With bounding box only & {28.5} & {-} & {-}\\

With mask only & {-} & {47.4} & {8.03} \\

\rowcolor{aliceblue} With both bounding box \& mask & \textbf{30.6} &\textbf{48.9} &\textbf{9.44} \\

\bottomrule
\end{tabular}
\vspace{-15pt}
\end{table}

\section{Experiment Results}
\label{sec:exp}

\subsection{Experimental Settings}
\label{subsec:experimental-settings}
\myheading{Datasets \& Evaluation Metrics.}
We conduct experiments on the Action Genome~\cite{ji2020action} for the DSGG task and the PVSG~\cite{yang2023pvsg} for the PVSG task. We evaluate the DSGG task following the setting from~\cite{Wang_2024_CVPR} and the PVSG task following the setting from~\cite{yang2023pvsg}. We train and evaluate the model strictly separately for each benchmark/task, without any data mixing or augmentation strategies.

\myheading{Implementation Details.}
We adopt both Vision Transformer (ViT) and Convolutional Neural Networks (CNNs) as the backbone for the Visual Encoder to perform frame-wise feature extraction. More specifically, for ViT, we use ViT-S/14, ViT-B/14, ViT-L/14~\cite{dosovitskiy2020vit} with pre-trained weights from DINO~\cite{oquab2023dinov2}; for CNNs, we use ResNet-50~\cite{DBLP:conf/cvpr/HeZRS16} pre-trained weights from MoCo~\cite{DBLP:conf/cvpr/He0WXG20}. 
{The number of slots are empirically chosen and can be observed that the optimal number of slots is influenced by both the number of object classes in the dataset and the objects present in the video.}
Therefore, for DSGG, we set $N=40$ object slots and $K=24$ relation slots; for PVSG, we use $N=96$ object slots and $K=40$ relation slots.

\subsection{Comparison with State of the Arts}
\myheading{Results on Box-Level VidSGG (DSGG).}
Tab.~\ref{tab:sota_dsgg} compares \model against SOTA methods on the Action Genome dataset.  
\model not only surpasses the second-best one-stage OED by a clear margin but also outperforms multi-stage methods with tracking mechanisms such as APT, TR\textsuperscript{2}, TPT, and TEMPURA.  
The results on the SGDET task highlight \model's strong VidSGG capabilities in DSGG.  
Our approach simultaneously localizes objects and predicts relations, achieving 45.2\% R@20 (\textuparrow 4.3\% over the second best) under \textit{With Constraint} and 49.7\% R@20 (\textuparrow 5.7\%) under \textit{No Constraint}.  
Similarly, in the PredCLS task, where oracle object tracks from ground truth are provided, \model surpasses all other methods across various metrics, reaching 80.3\% R@20 under \textit{With Constraint} and 98.1\% R@20 under \textit{No Constraint}.  
Despite multi-stage methods benefiting from oracle tracks and directly aggregating accurate spatio-temporal context, \model still outperforms them.  
However, our results indicate room for improvement in object localization.


\myheading{Results on Pixel-Level VidSGG (PVSG).}
The PVSG dataset presents highly dynamic videos and frequent substantial changes in camera angles. Tab.~\ref{tab:pvsg_sota} demonstrates that both the Image Panoptic Segmentation + Tracking~\cite{cheng2021mask2former,wang2021different} (IPS+T) model and Video Panoptic Segmentation~\cite{cheng2021mask2former,li2022videoknet} (VPS) baselines fall short compared to our end-to-end \model.
It is essential to focus on R/mR@20, as it serves as a primary performance metric~\cite{yang2023pvsg}. Notably, our model substantially outperforms both IPS+T and VPS at R/mR@20 across mask overlaps of 0.5 and 0.1. For instance, \model holds the current peak for R@20 at 9.44\% in the vIOU threshold of 0.5, indicating that, on average, 01 in every 13 ground-truth triplets is successfully recalled, compared to IPS+T's best performance of 01 in roughly every 25 triplets (R@20 of 4.51\%). However, lowering the threshold to a more lenient 0.1 raises \model's score to approximately 17.54\%, allowing the model to recall 02 out of every 13 triplets. This suggests that while the model shows higher efficacy than others for recognizing key video content, there remains considerable room for improvement.

\subsection{Ablation Studies}
\label{subsec:ablation}

\myheading{Effect of Multi-Task Learning.} Tab.~\ref{tab:multi_task} highlights the impact of multi-task learning and cross-task synergy on PVSG, which provides both box-level and pixel-level VidSGG ground truths. AP50 denotes Average Precision@0.5, while PQ is the Panoptic Quality~\cite{kirillov2019panoptic}. Using \model, we observe that training with both bounding boxes and masks improves all metrics, with AP50 increasing from 28.5 to 29.6, PQ from 47.4 to 48.3, and R@20 from 8.03 to 9.44. These results suggest that box-level and pixel-level VidSGG data complement each other, enabling richer representations that enhance individual task performances. 

\begin{table}[!t]
\footnotesize
\centering
\caption{{Temporal consistency learning ablation.}}
\vspace{-5pt}
\label{tab:temporal}
 \setlength{\tabcolsep}{2pt}
\resizebox{0.9\linewidth}{!}{
\begin{tabular}{lcccccc}
\toprule
\multicolumn{1}{c}{\multirow{2}{*}{\textbf{Method}}} 
& \multicolumn{3}{c}{\textbf{DSGG task}} 
& \multicolumn{3}{c}{\textbf{PVSG task}} \\ 
\cmidrule(lr){2-4}\cmidrule(lr){5-7}
& {R@10$\uparrow$} & {R@20$\uparrow$} & \multicolumn{1}{c}{R@50$\uparrow$}  & {R@20$\uparrow$} & {R@50$\uparrow$} & \multicolumn{1}{c}{R@100$\uparrow$} \\ 
\midrule

\text{w/o $\mathcal{L}_\text{consistency}$} & {38.1} & {41.3} & {49.4} & {7.16} & {9.23} & {12.17} \\

\rowcolor{aliceblue} \text{w/ $\mathcal{L}_\text{consistency}$} & \textbf{39.3} &\textbf{45.2} &\textbf{53.8} & \textbf{9.44} & \textbf{10.83} & \textbf{11.59} \\

\bottomrule
\end{tabular}
}
\vspace{-20pt}
\end{table}

\myheading{Multi-Object Spatio-Temporal Consistency.} By enabling \model with slot attention to capture modular object features from frozen visual representations, we detect a form of temporal consistency, where object slots retain stable representations as a video sequence evolves, which can be observed from Tab.~\ref{tab:temporal} across both tasks even without $\mathcal{L}_\text{consistency}$. By explicitly incorporating $\mathcal{L}_\text{consistency}$ that aims to align slots through time, we observe a significant improvement. It is also supported from the results of PVSG in Tab.~\ref{tab:pvsg_sota}, where volume IOU is involved to consider mask consistency through time. Such finding results in what we term Multi-Object Spatio-Temporal Consistency, where spatial features (slots binding to visual features) and temporal features (object transitions over time) are cohesively integrated for improved accuracy via \model, with one such case illustrated in Fig.~\ref{fig:quali_temporal}.


\begin{figure}[t]
    \centering
    \includegraphics[width=\linewidth]{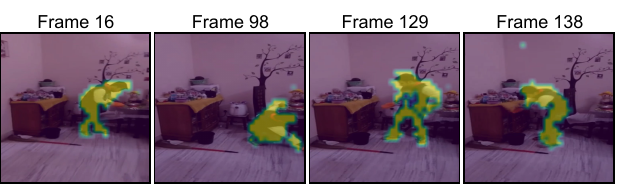}
    \vspace{-2em}
    \caption{Spatio-temporal consistency of an object slot over time.}
\label{fig:quali_temporal}
\vspace{-10pt}
\end{figure}

\begin{figure}[t]
    \centering
    \includegraphics[width=\linewidth]{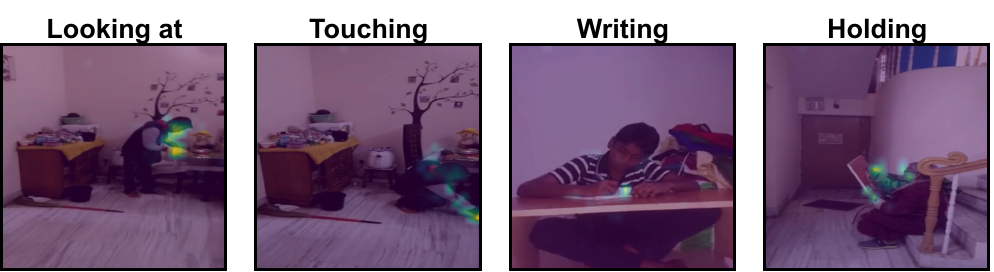}
    \vspace{-2em}
    \caption{Visualization results of relation slots on Action Genome.}
\label{fig:relation_slot_viz_main}
\vspace{-10pt}
\end{figure}

\begin{figure}[!t]
    \centering
    \includegraphics[width=0.9\linewidth]{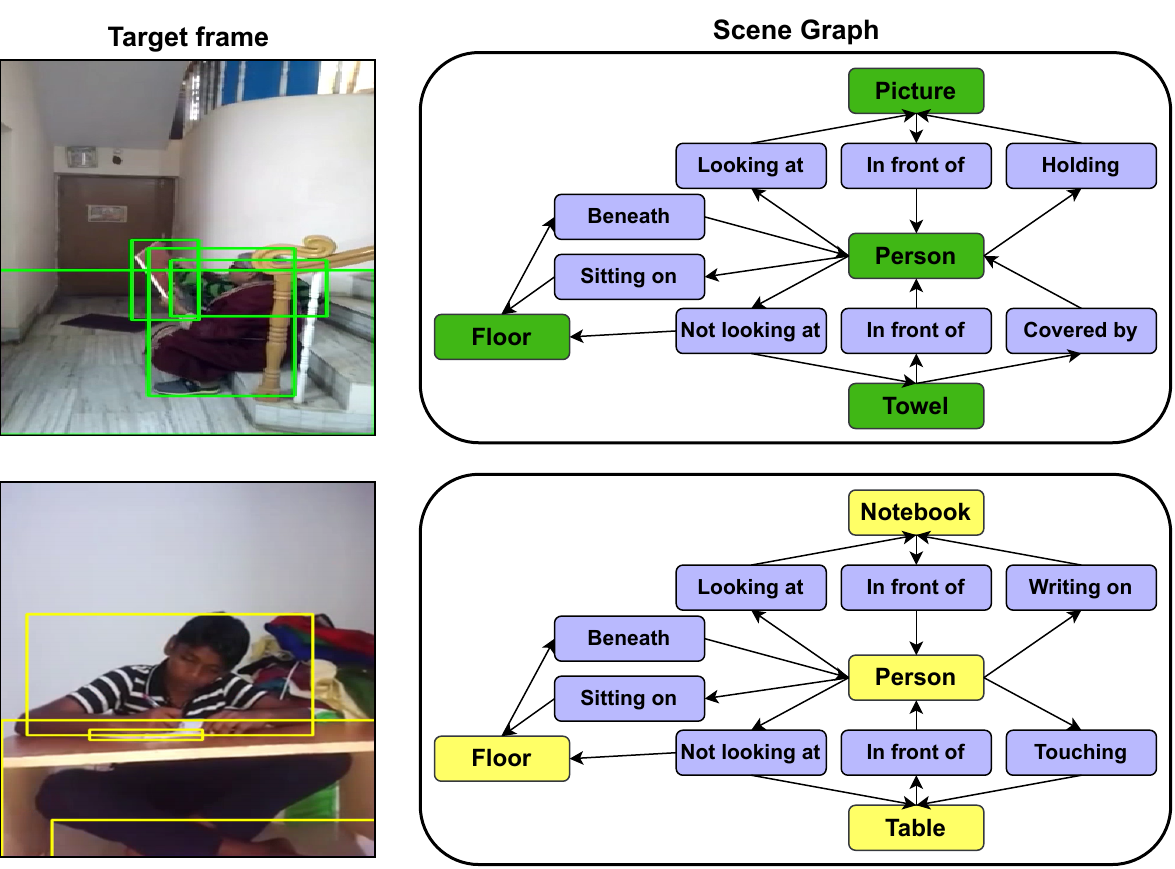}
    \vspace{-1em}
    \caption{Visualization results  on DSGG task. we illustrate a case of a ``Person" sitting on the ``Floor" while holding a ``Picture", with a ``Towel" wrapped around his back. }
\label{fig:dsgg_viz_1}
\vspace{-10pt}
\end{figure}

\begin{figure}[t]
    \centering
    \includegraphics[width=\linewidth]{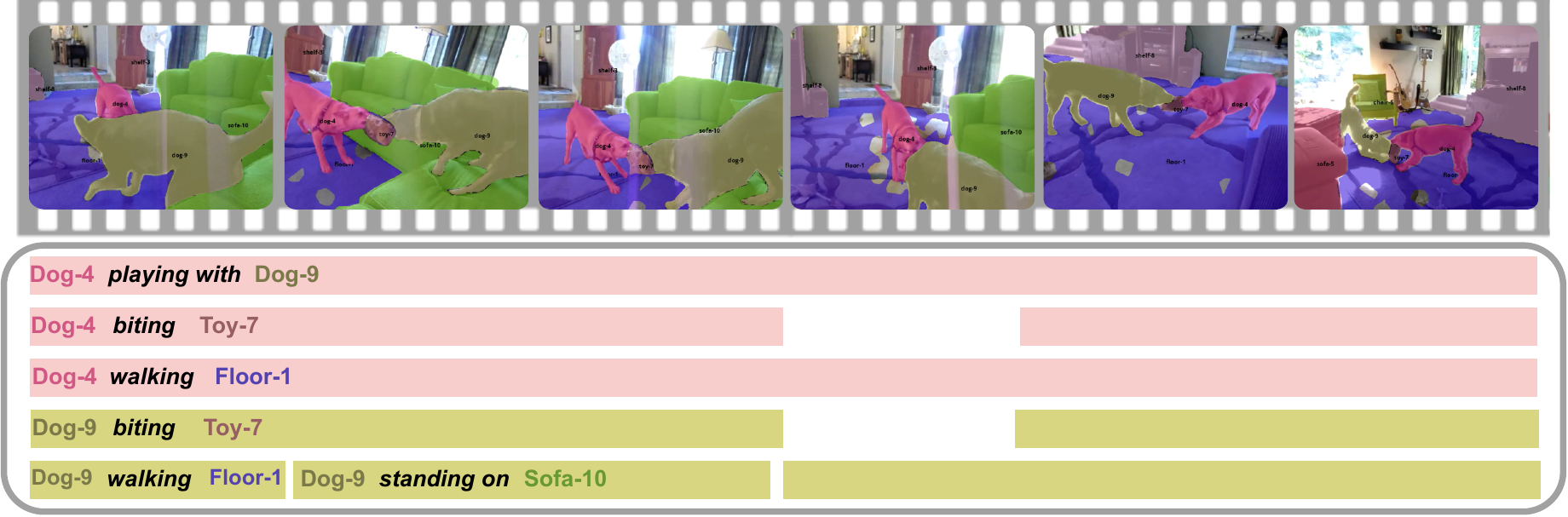}
    \vspace{-2em}
    \caption{Visualization results of \model on PVSG. \model is qualitatively shown to address a complex task with a video of two dogs (\ie ``Dog-4" and ``Dog-9") playing around in a living room with a toy (\ie ``Toy-7") on the floor (\ie ``Floor-1") and the sofa (\ie ``Sofa-10"), cluttered in the background are miscellaneous objects (\ie ``Shelf-3", ``Shelf-8").}
\label{fig:PVSG_viz_1}
\vspace{-15pt}
\end{figure}

\subsection{Qualitative Results}
\myheading{DSGG.}
Fig.~\ref{fig:quali_temporal} illustrates a long time step (frame 16 to frame 138) of a test video sequence in Action Genome. It reveals that our method is able to distinguish object instances with semantic structure at the mask level through object slots, even without such labels in Action Genome. This indicates that \model is able to maintain a coherent spatio-temporal representation. 
Fig.~\ref{fig:relation_slot_viz_main} visualize relation slots of a test video sequence in Action Genome. It demonstrates the ability of \model to capture meaningful relational semantics via slot attention, which are interestingly shown as highlighted areas between actors and objects, indicating that \model can interpret structural semantics that correspond with relations in a spatial manner. Another example is shown in Fig.~\ref{fig:dsgg_viz_1}, where \model can capture the person, objects, and their interactions.

\myheading{PVSG.} {Fig.~\ref{fig:PVSG_viz_1} visualizes the result of \model on PVSG. On the top part, \model demonstrates the extraction of semantic masks from an example video, and in the bottom part, it showcases the consistent prediction of object-relation semantics across time. This emphasizes \model's ability to handle dynamic interactions and complex environments.}

\vspace{-0.5em}
\section{Conclusion}
\vspace{-0.5em}
\label{sec:conclusion}
We propose \model, a unified framework that effectively addresses both coarse-grained and fine-grained tasks. By incorporating an enhanced slot attention mechanism and object temporal consistency learning, \model learns robust, modular representations that adapt dynamically to box-level and pixel-level visual granularity. Additionally, we integrate a dynamic triplet prediction module to establish precise, relation-specific associations between objects, improving efficiency while reducing redundancy. Our empirical results on Action Genome and PVSG convey that \model offers an effective, streamlined solution for VidSGG, advancing the state-of-the-art in unified, object-centric spatio-temporal representation learning for VidSGG.

\myheading{Limitations.}
\model may face challenges in handling object disappearances or reappearances. It also operates with a fixed number of slots that limit its adaptability to complex dynamics, such as crowded environments, fast-motion scenarios, or when numerous small objects quickly maneuver.

\myheading{Broader Impacts.}
\model introduces a new paradigm for unified VidSGG, providing a computationally efficient and flexible framework that facilitates easy development and serves as a structured representation generation to enhance a wide range of video understanding and reasoning tasks.
\clearpage
\newpage
{
    \small
    \bibliographystyle{ieeenat_fullname}
    \bibliography{main}
}

\end{document}